\documentclass[letterpaper]{article} 
\usepackage{xai4sci}
\usepackage{times}  
\usepackage{helvet}  
\usepackage{courier}  
\usepackage[hyphens]{url}  
\usepackage{graphicx} 
\usepackage{natbib}  
\usepackage{caption} 
\usepackage{subcaption}
\usepackage[ruled,vlined,linesnumbered]{algorithm2e}
\SetKwComment{Comment}{$\triangleright$\ }{}
\SetKw{KwEach}{each}
\SetKw{KwIn}{Input:}
\SetKw{KwOut}{Output:}
\SetKw{Kwin}{in}
\SetKw{KwBreak}{break}

\usepackage{xcolor}
\usepackage{amsmath,amsfonts,amsthm,amssymb,upgreek}

\title{Shapelet-based Model-agnostic Counterfactual Local Explanations for Time Series Classification}
\author {
    Qi Huang\thanks{These authors contribute equally to the paper} \thanks{Corresponding author}, Wei Chen\footnotemark[1],
    Thomas B{\"a}ck,
    Niki van Stein
}
\affiliations {
    LIACS, Leiden University\\
    Niels Bohrweg 1, 2333 CA, Leiden \\
    The Netherlands\\
    q.huang@liacs.leidenuniv.nl
}

\usepackage{bibentry}

\begin{document}

\maketitle

\begin{abstract}


In this work, we propose a model-agnostic instance-based post-hoc explainability method for time series classification. The proposed algorithm, namely \textbf{Time-CF}, leverages \textit{shapelets} and \textit{TimeGAN} to provide counterfactual explanations for arbitrary time series classifiers. We validate the proposed method on several real-world univariate time series classification tasks from the \textit{UCR Time Series Archive}. The results indicate that the counterfactual instances generated by \textbf{Time-CF} when compared to state-of-the-art methods, demonstrate better performance in terms of four explainability metrics: closeness, sensibility, plausibility, and sparsity.

\end{abstract}

\section{Introduction}
In recent years, the performance of machine learning models has become increasingly impressive. Advances in the capability of data computation, the development of sophisticated algorithms, and the proliferation of high-quality datasets have all contributed to this upward trend \cite{dl_advantage,ml_advantage}.
Machine learning models and methods are being leveraged in various aspects of daily life and science, such as automated driving, banking, healthcare, physics, and geoscience, where intelligible explanations for their decision-making processes are essential for establishing public trust over utilized opaque models \cite{need_XAI}. In addition, these methods can also generate new insights for researchers in various domains.
Especially in safety-critical applications such as medical diagnosis and self-driving cars, mistakes in predictions could be fatal; thus, there is an urgent need either to provide \textbf{post-hoc} explanations based on the decisions made by the models or to build inherently explainable models before predictions \textbf{(ante-hoc)}.

In response to the opacity property of machine learning models, the field of eXplainable AI (XAI) has emerged to address this issue. Although XAI methods have been proposed and effectively applied to \textit{static} data such as images, text, and tabular data, there has been relatively less work on time series data \cite{need_focus_on_ts_data}.
The existing XAI approaches for time series analysis can be categorized into three types regarding the level of explanation: point-level, subsequence-level, and instance-level.
Among the existing methods, counterfactual explanations for instance-level have recently gained wider popularity \cite{xai_review0}.

As its name suggests, an instance-level counterfactual explanation for a to-be-explained (target) time series is an artificially generated instance, derived from the target, that does not exist in original observations but contradicts the original facts e.g., predicted to be a different class in classification.
Perturbing the target time series, where minimal changes are preferred, is one of the most common techniques for creating counterfactual explanations for time series instances~\cite{SG-CF,MG-CF}. Explainability is hence achieved by highlighting the difference between the original instance and its generated counterpart.
The focus of this research is to use a Generative Adversarial Network (GAN) to fabricate counterfactual explanations that can explain the predictive behavior of time series classifiers on an instance level.

\begin{figure*}[tb!]
    \centering
\includegraphics[width=0.8\textwidth,height=0.22\textheight]{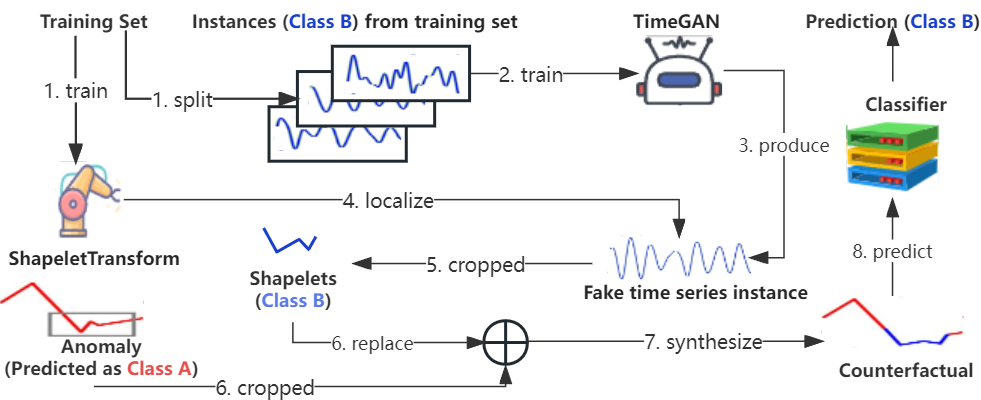}
    \caption{The flowchart of Time-CF in binary classification. The process describes how the counterfactual explanation of an instance labeled as \textcolor{red}{Class A}, is synthesized by TimeGAN and shapelet transform, based on instances from \textcolor{blue}{Class B}.}
    \label{fig:cf_model}
\end{figure*}

\section{Related Works}
\label{chapter:related_work}
In this work, we focus on counterfactual-based XAI methods for Time Series Classification (TSC). TSC stands for a type of machine learning task where a machine learning model (classifier) shall be learned to maximally correctly assign labels to a set of time series instances. We refer the readers to the survey paper regarding XAI for time series classification~\cite{xai_review0} for additional information.

Wachter \textit{et al.} are pioneers in generating counterfactual instances to elucidate anomalies \cite{w-CF}. This involves simultaneously minimizing a prediction loss function that drives the prediction on counterfactuals towards a different class and a distance loss function that ensures the similarity between counterfactuals and to-be-explained instances. Recent efforts to use counterfactual instances as explanations are mostly derived from this work.
\textbf{Native-Guide} \cite{NativeGuide} is another counterfactual-based method, introduced by \textit{Delaney et al}, following four desirable properties: Sparsity, Proximity, Plausibility, and Diversity. In essence, they find the nearest-unlike neighbor of the to-be-explained instance and generate a counterfactual instance for explanation.
\textbf{MG-CF} \cite{MG-CF}, \textbf{SG-CF} \cite{SG-CF}
are also extensions of \cite{w-CF}, but incorporate shapelets into consideration. Specifically, \textbf{MG-CF} generates counterfactual instances by using \textbf{shapelet Transform} \cite{shapeletTransform} to extract all potential shapelet candidates and then replace the same part of the to-be-explained instance. Moreover, \textbf{SG-CF} adds an additional term to the loss function used in \cite{w-CF}, which ensures high Euclidean-based similarity between the counterfactual instance and the shapelet, thus providing more robust explanations.
Besides directly perturbing the original sequence which is essentially injecting random noise to the region-of-interest, \cite{SPARCE} and \cite{satellite-GAN} propose to employ the generative adversarial networks~\cite{GAN} to \textit{fabricate} entire counterfactual instances, which also enlightens us in developing our method.

\section{Methodology}
\label{chapter:methodology}
\begin{figure}[tb]
    \centering
\includegraphics[width=0.4\textwidth, height=0.15\textheight]{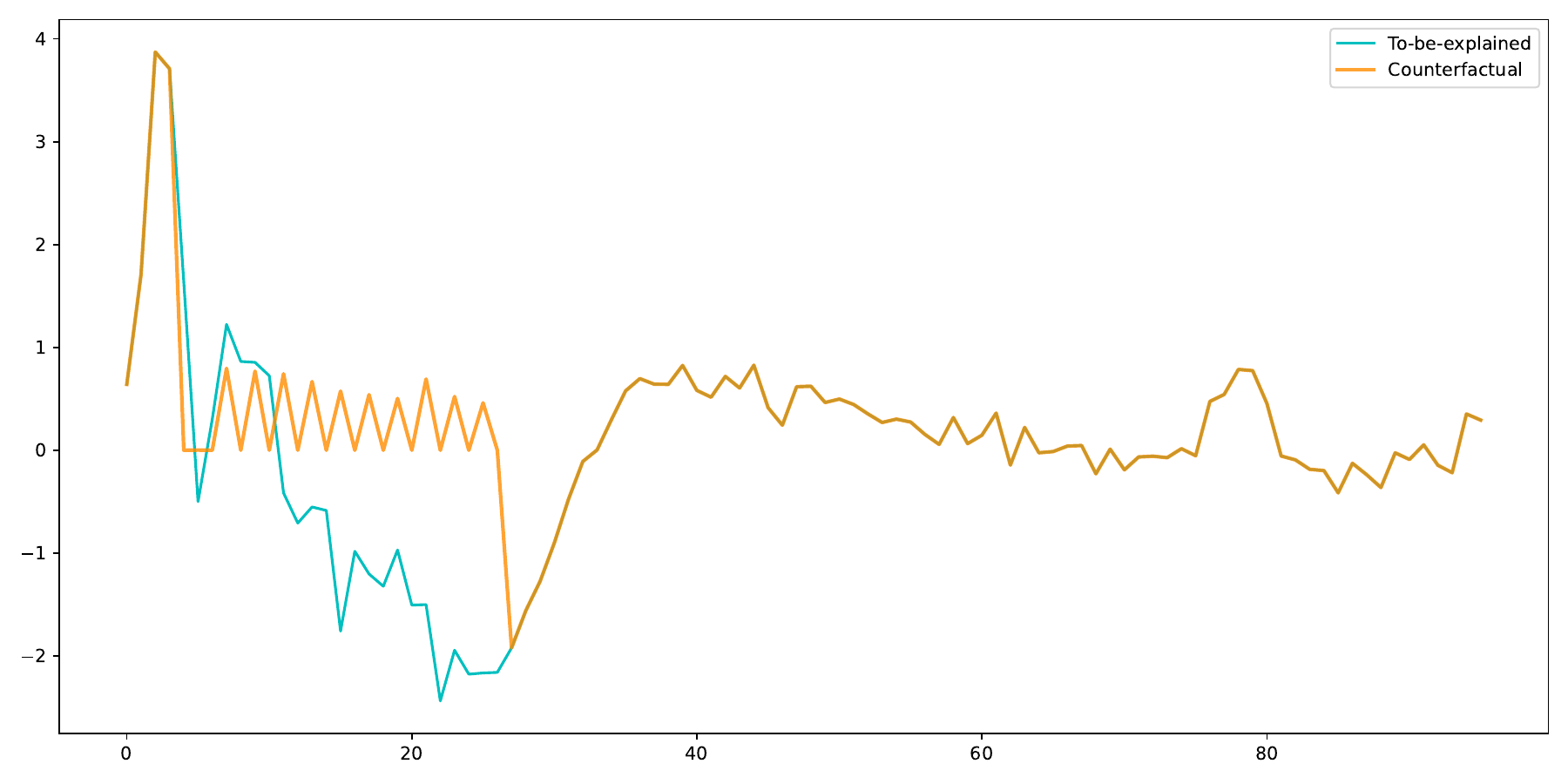}
    \caption{An example of a time series counterfactual instance, where the blue curve represents the to-be-explained shapelet of the electrocardiogram instance from the UCR ECG200~\cite{UCRArchive}. The orange curve stands for its generated counterfactual subsequence, which drives the classifier to alter its prediction.}
    \label{fig:cf_example}
\end{figure}

Having discussed related work on counterfactual-based instance-level XAI for time series classification, we now introduce our proposed algorithm, \textbf{Time-CF}. A flowchart of the algorithm under binary classification is shown in Figure \ref{fig:cf_model}. An example outcome of Time-CF is displayed in Figure~\ref{fig:cf_example}, where only a subsequence (the blue curve) of the to-be-explained instance is altered (to the orange curve).

The process starts with extracting a designated number of shapelet candidates of various lengths from the time series instances.
Shapelets are subsequences of a time series instance that can be extracted by shapelet transform~\cite{shapeletTransform,shapelet_transform_tsc_2}.
When it comes to instance-level explanations, shapelets can be directly projected onto the original time series, thus providing more informative and intuitive explanations that require minimal domain knowledge. A typical Shapelet Transform (ST) takes as input the training set and outputs extracted shapelets. The transform is composed of three steps: candidate extraction, computation of information gain per candidate, and shapelet selection. As a consequence of the selection, shapelets with less discriminative information (w.r.t. class labels) are filtered out, and those with relatively high discriminative power are retained for the perturbation steps. In this study, we utilize the Random Shapelet Transform (RST), which accommodates time limitations and random candidate extraction, thus providing time-saving and flexibility advantages in comparison to the complete shapelet transform.
After shapelet extraction, the collected candidates are sorted by their respective information gain in descending order and then filtered to the top N (the number of retained shapelets).
Inspired by the uprising interest in utilizing generative models for XAI in time series analysis \cite{conditional_gan_for_all, SPARCE, satellite-GAN}, we likewise alternatively apply generative adversarial networks dedicated for time series data namely, \textit{TimeGAN}~\cite{TimeGAN}, as our counterfactual generation model. 



\begin{algorithm*}[hbt!]
\caption{Generating counterfactual explanations of a classifier $f$ for a time series instance}\label{alg:generating_cf}
\KwData{shapelet candidates: $S$, a to-be-explained instance with class label \text{$l$}: $T$, the length of $T$: $N$, a to-be-explained classifier $f$, the maximum number of generated sequences: $M$}

\KwResult{a collection of counterfactual instances: $C$}

$C \gets \emptyset$ \Comment*[r]{A set of counterfactual instances for $T$}
$F \gets \emptyset$ \Comment*[r]{A set of generated \textit{fake} instances}
\text{Train a TimeGAN with instances of other labels\;}\\

\For{$i \leftarrow 1$ \KwTo $M$}{
  $\mathbf{x} \gets \left[ x_1, x_2, \ldots, x_N \right]^T \sim \text{Uniform}[0, 1]$ \Comment*[r]{Sample the initial vector}
  $T_{fake} \gets \text{TimeGAN}(x)$ \Comment*[r]{Transform $x$ to a \textit{fake} sequence}
  $F \gets F \cup \{T_{fake}\}$
}
\For{$s$ in $S$}{
    $\text{start\_pos}, s_{\text{len}} \gets \text{s.info}$ \Comment*[r]{Get the starting index and length of $s$}
  \If{s.class == l}
  {
    \Comment*[f]{Retrieve shapelets that belong to the same interval as $s$ from each $T_{fake}$ in $F$}\\
    $S_{fake} \gets \text{crop}(\text{start\_pos}, s_{\text{len}}, $F$)$\;
    
    \For{$s_j$ in $S_{fake}$}{
    $T_{fake} \gets \text{replace}(T, s_{j}, \text{start\_pos}, s_{\text{len}})$ \Comment*[r]{Replace the target subsequence in $T$ with $s_{j}$}
    \If{$f(T_{fake}) \neq s.class$}
    {
    $C \gets C \cup \{T_{fake}\}$
    }}
  }
}
  \Return $C$
\end{algorithm*}
Algorithm \ref{alg:generating_cf} depicts how the counterfactual instances are generated. With a given to-be-explained time series instance $T$ with class label $l$, we first train a TimeGAN using all instances from the training data except for those with label $l$.
The trained TimeGAN is then used to generate artificial instances. The final output sequences of TimeGAN can closely resemble the sequence-level distribution of the training data while preserving the within-sequence stepwise transition~\cite{TimeGAN}. As depicted in lines 5 to 7 of the pseudo-code, a total of $M$ fake instances, each of length $N$, are obtained based on $M$ initial vectors. These initial vectors are uniformly sampled from $[0,1]^{N}$.
Next, the algorithm iterates over the shapelet candidate set $S$. For each candidate $s$, it is essential to identify its start and end positions within the corresponding instance.
Further, the $crop()$ function is executed to extract the artificial shapelets $S_{fake}$ by cropping each interval with the same time range (start and end positions) in each generated fake instance $T_{fake}$ from $F$, that is, an instance from the same class. The final process includes iterating through the set of fake shapelets and using the $replace()$ method to replace the contiguous time points (ranging from the start position and end position stored previously) of the to-be-explained time series instance $T$, with one of the (fake) shapelets generated by TimeGAN. A synthetic time series instance $T_{fake}$ is considered as a counterfactual only if it flips the prediction of the to-be-explained classifier, says, the predictive label of $T_{fake}$ is no longer $l$. Moreover, this step reserves all the fake counterfactual time series instances for diverse explanations. In case of finding multiple counterfactual instances, the final recommended one is determined by:
\begin{equation} \label{equation:w-cf}
T_c \gets \underset{\  \ \ \ T_{f} \in C}{\text{argmin}}\ d(T_{f}, T)
\end{equation}
where $d$ is a (distance) metric defined over the given time series. Specifically, we minimize the Hamming distance in this study to ensure minimum perturbation actions.

\section{Experimental Setup}
Time-Cf is evaluated on four types of classifiers (Convolutional Neural Networks (CNNs), Diverse Representation Canonical Interval Forest (DrCIF), Catch22 with Random Forest (RF), and K Nearest Neighbor (KNN) Classifier. Each classifier is trained with four binary classification time series data from the UCR time series classification archive~\cite{UCRArchive}. They are ECG200, an electrocardiogram; Wafer, sensor signals during silicon wafer processing for semiconductor fabrication; FordA, readings of engine noise for outlier detection; MoteStrain, sensor measurements of humidity or temperature. The results obtained by Time-CF on these four datasets are compared with those of Native Guide~\cite{NativeGuide} and Mlxtend~\cite{mlxtend}, where the former is a well-established counterfactual-based method (see related works) and the latter is an open-source library tailored for interpreting machine learning classifiers.

\section{Analysis of Results}

Although there is no consensus on the evaluation of counterfactual-based methods, closeness, sparsity, plausibility, and diversity have been widely used in recent studies
~\cite{NativeGuide, four-measures-0, MG-CF, SG-CF}.
In a similar vein, we utilize the first three measures and replace diversity with sensibility in our evaluation since the existence of instance-wise explanations is more crucial and questionable, as observed in our study (see Fig.~\ref{fig:experiment_omission_study}).


\subsubsection{Closeness}
\begin{figure}
\centering
\includegraphics[width=0.45\textwidth, height=0.2\textheight]{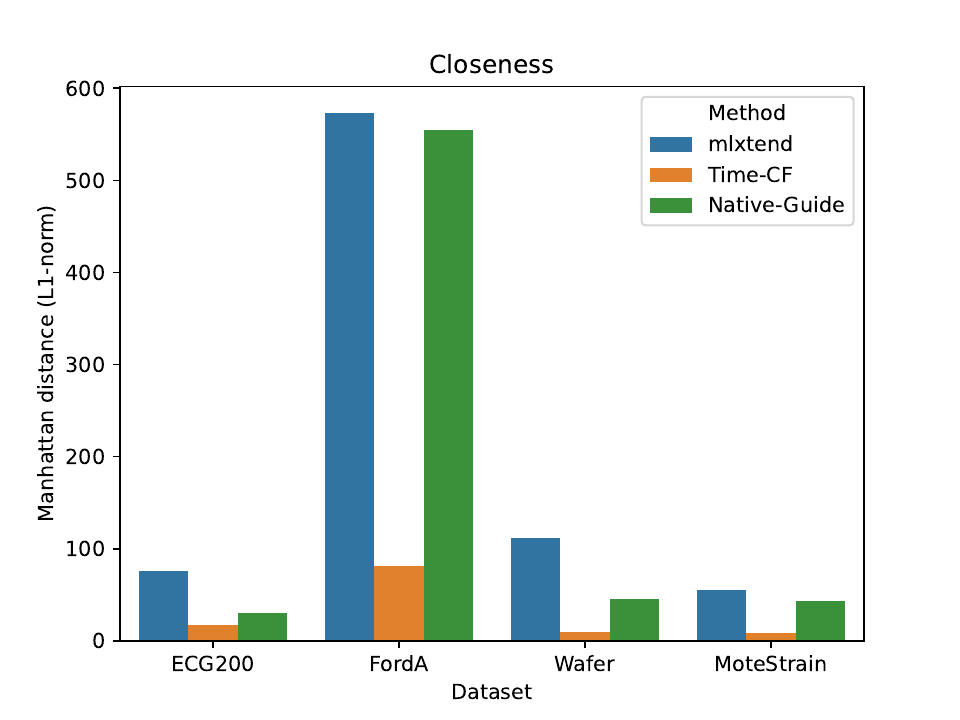}
\caption{The results of L1-norm closeness. A low value is preferred, as it indicates less scale change.}
\label{fig:closeness_comparative_study}
\end{figure}
It measures the \textit{shape} similarity between the counterfactual instance and the original instance in Manhattan distances~\cite{closeness-measure-0}. Figure \ref{fig:closeness_comparative_study} shows that the counterfactual instances made by our method are the closest to the instances being explained in all datasets. This reveals Time-CF's proficiency in perturbing the minimal, contiguous segment of the original instances.

\subsubsection{Sensibility}
\begin{figure}
    \centering
\includegraphics[width=0.5\textwidth, height=0.15\textheight]{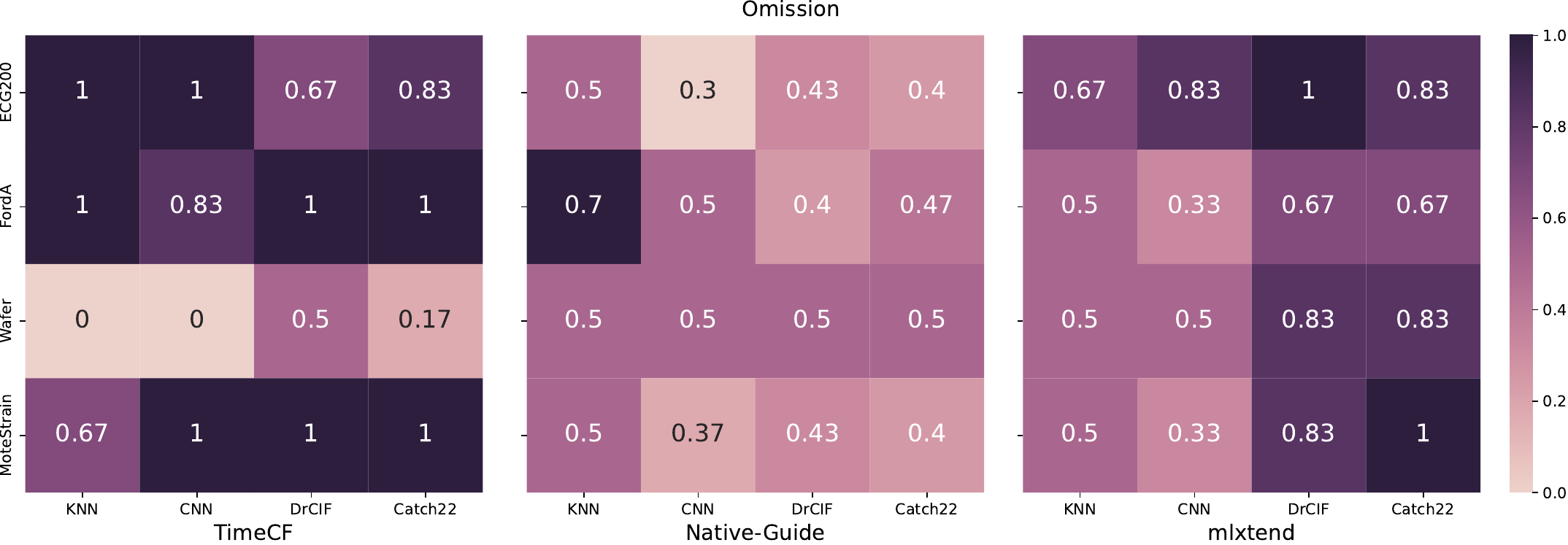}
    \caption{Assessment of sensibility, where \textbf{higher} values (darker in plot) are preferable. The figure displays the ratios of each post-hoc explainer in \textbf{successfully} finding counterfactual instances across four datasets.
    }
    \label{fig:experiment_omission_study}
\end{figure}

Sensibility serves as a metric to determine the extent to which counterfactual-based methodologies are model-agnostic or can explain any classifier. It records the proportion of instances per classifier that are explained by the given XAI approach.
The results in Fig. \ref{fig:experiment_omission_study} demonstrate that both \textbf{mlxtend} and \textbf{Native-Guide} fail to effectively handle a variety of classifiers across these datasets. In contrast, with the exception of the extremely imbalanced dataset \textit{Wafer}, our method can aptly discern the informative features in time series data, thereby positioning itself as a model-agnostic tool that can, in theory, clarify any classifier.

\subsubsection{Plausibility}

\begin{figure}
    \centering
\includegraphics[width=0.45\textwidth, height=0.25\textheight]{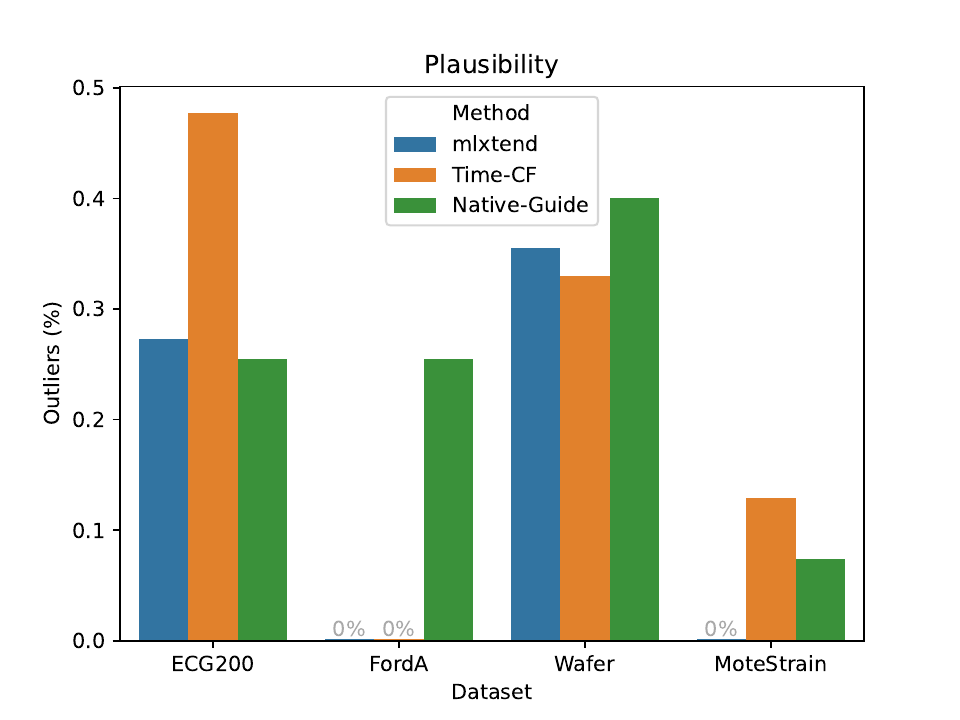}
    \caption{
    Assessment of plausibility, where a lower value is preferable. The y-axis quantifies the outlier ratios of the generated counterfactual instances.
    }
    \label{fig:plausibility_comparative_study}
\end{figure}

In this experiment, we delve deep into the percentage of generated instances identified as out of distribution. Given our emphasis on the importance of plausibility when generating counterfactual instances, we exclusively retain those counterfactuals with the lowest outlier rates to ensure heightened plausibility.  As depicted in Fig. \ref{fig:plausibility_comparative_study}, our method outperforms others in terms of plausibility in dataset Wafer. However, it shows worse performance than the others when using an imbalanced dataset. Note that this metric is dependent on the outlier detection algorithm, which in this study is the Isolation Forest.

\subsubsection{Sparsity}

\begin{figure}
    \centering
\includegraphics[width=0.48\textwidth, height=0.2\textheight]{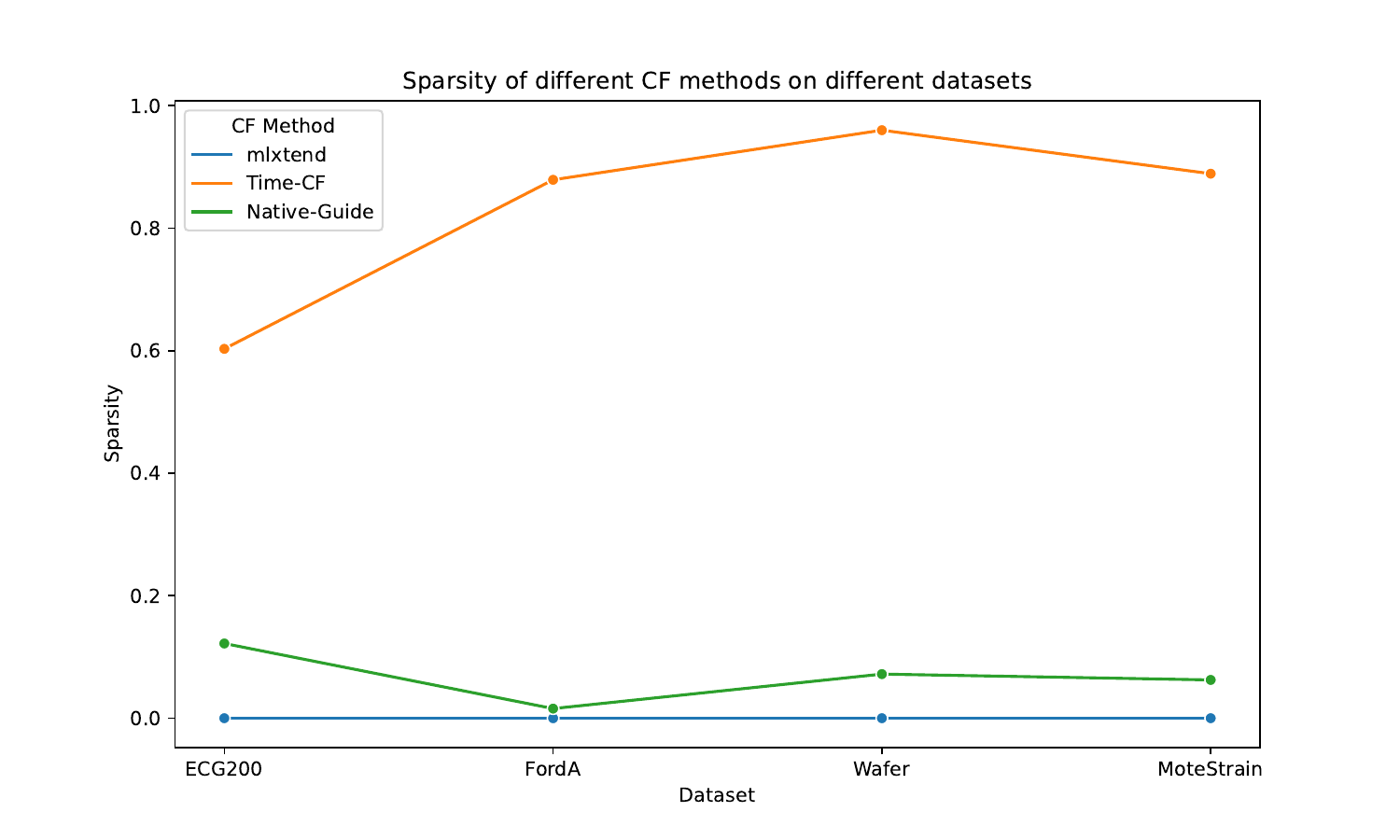}
    \caption{Measurement for sparsity (the higher, the better). This metric indicates the extent to which the number of time steps is altered. A value of 0 means that the counterfactual instances are derived from the original instances that have alterations at every time point.}
    \label{fig:sparsity_comparative_study}
\end{figure}

Lastly, 
curve graph in Fig. \ref{fig:sparsity_comparative_study} contrasts the sparsity level of each XAI method. When comparing the sparsity levels among \textbf{mlxtend} (blue), \textbf{Native-Guide} (green) and Time-CF (yellow), our approach consistently showcases superior performance. It is evident that both \textbf{mlxtend} and \textbf{Native-Guide} require modifications to nearly all the time steps to alter the class of the to-be-explained instance from one to another, with values approaching a sparsity of 0\%. In contrast, our approach only perturbs a minor segment, thereby offering not only clearer visualization for analytical objectives but also a more intuitive comprehension of which portions primarily influence the classifier.

\section{Conclusion}
\label{chapter:conclusion}

In this work, a model-agnostic, counterfactual-based XAI method, \textbf{Time-CF} is introduced to explain any prediction for any classifier concerning time series classification tasks. \textbf{Time-CF} chiefly utilizes shapelets and TimeGAN to deliver meaningful counterfactual explanations, exhibiting exceptional performance in terms of closeness, sensibility, plausibility, and sparsity measures. The incorporation of shapelets indicates that a contiguous segment of the time series instance is perturbed, thus offering intuitive insights for the analysis of anomalies for end-users. In addition, TimeGAN ensures that at least one counterfactual explanation is generated, provided the training set is both ample and not extremely imbalanced. TimeGAN also guarantees that the explanations produced adhere to the data distribution.

In future work, we aim to refine our methodology to cater to multi-class and multivariate time series classification tasks. Moreover, a more profound exploration of the synergies between TimeGAN and Shapelet Transform (ST) will be undertaken to address the suboptimal performance observed with imbalanced datasets, and a wider set of experiments will be conducted to assess the behavior and performance of the proposed method against other purely GAN-based methods~\cite{SPARSE_GAN}.

\section{Acknowledgement}
This publication is part of the project XAIPre (with project number 19455) of the research program Smart Industry 2020 which is (partly) financed by the Dutch Research Council (NWO).
\bibliography{submission-aaai.bib}

\begin{thebibliography}{21}
\providecommand{\natexlab}[1]{#1}

\bibitem[{Bostrom and Bagnall(2017)}]{shapelet_transform_tsc_2}
Bostrom, A.; and Bagnall, A. 2017.
\newblock Binary Shapelet Transform for Multiclass Time Series Classification.
\newblock In Hameurlain, A.; K{\"u}ng, J.; Wagner, R.; Madria, S.; and Hara, T., eds., \emph{Transactions on Large-Scale Data- and Knowledge-Centered Systems {{XXXII}}: {{Special}} Issue on Big Data Analytics and Knowledge Discovery}, 24--46. {Berlin, Heidelberg}: {Springer Berlin Heidelberg}.
\newblock ISBN 978-3-662-55608-5.

\bibitem[{Chen et~al.(2015)Chen, Keogh, Hu, Begum, Bagnall, Mueen, and Batista}]{UCRArchive}
Chen, Y.; Keogh, E.; Hu, B.; Begum, N.; Bagnall, A.; Mueen, A.; and Batista, G. 2015.
\newblock The UCR Time Series Classification Archive.
\newblock \url{www.cs.ucr.edu/~eamonn/time_series_data/}.

\bibitem[{Dantas, Marcos, and Ienco(2023)}]{satellite-GAN}
Dantas, C.~F.; Marcos, D.; and Ienco, D. 2023.
\newblock Counterfactual Explanations for Land Cover Mapping in a Multi-class Setting.
\newblock \emph{arXiv preprint arXiv:2301.01520}.

\bibitem[{Delaney, Greene, and Keane(2021{\natexlab{a}})}]{NativeGuide}
Delaney, E.; Greene, D.; and Keane, M.~T. 2021{\natexlab{a}}.
\newblock Instance-based counterfactual explanations for time series classification.
\newblock In \emph{Case-Based Reasoning Research and Development: 29th International Conference, ICCBR 2021, Salamanca, Spain, September 13--16, 2021, Proceedings 29}, 32--47. Springer.

\bibitem[{Delaney, Greene, and Keane(2021{\natexlab{b}})}]{closeness-measure-0}
Delaney, E.; Greene, D.; and Keane, M.~T. 2021{\natexlab{b}}.
\newblock Uncertainty estimation and out-of-distribution detection for counterfactual explanations: Pitfalls and solutions.
\newblock \emph{arXiv preprint arXiv:2107.09734}.

\bibitem[{Goodfellow et~al.(2020)Goodfellow, Pouget-Abadie, Mirza, Xu, Warde-Farley, Ozair, Courville, and Bengio}]{GAN}
Goodfellow, I.; Pouget-Abadie, J.; Mirza, M.; Xu, B.; Warde-Farley, D.; Ozair, S.; Courville, A.; and Bengio, Y. 2020.
\newblock Generative adversarial networks.
\newblock \emph{Communications of the ACM}, 63(11): 139--144.

\bibitem[{Hills et~al.(2014)Hills, Lines, Baranauskas, Mapp, and Bagnall}]{shapeletTransform}
Hills, J.; Lines, J.; Baranauskas, E.; Mapp, J.; and Bagnall, A. 2014.
\newblock Classification of time series by shapelet transformation.
\newblock \emph{Data mining and knowledge discovery}, 28: 851--881.

\bibitem[{Lang et~al.(2022)Lang, Giese, Ilg, and Otte}]{SPARCE}
Lang, J.; Giese, M.; Ilg, W.; and Otte, S. 2022.
\newblock Generating Sparse Counterfactual Explanations For Multivariate Time Series.
\newblock \emph{arXiv preprint arXiv:2206.00931}.

\bibitem[{Lang et~al.(2023)Lang, Giese, Ilg, and Otte}]{SPARSE_GAN}
Lang, J.; Giese, M.~A.; Ilg, W.; and Otte, S. 2023.
\newblock Generating Sparse Counterfactual Explanations for Multivariate Time Series.
\newblock In Iliadis, L.; Papaleonidas, A.; Angelov, P.; and Jayne, C., eds., \emph{Artificial Neural Networks and Machine Learning -- ICANN 2023}, 180--193. Cham: Springer Nature Switzerland.
\newblock ISBN 978-3-031-44223-0.

\bibitem[{Li et~al.(2022)Li, Bahri, Boubrahimi, and Hamdi}]{SG-CF}
Li, P.; Bahri, O.; Boubrahimi, S.~F.; and Hamdi, S.~M. 2022.
\newblock SG-CF: Shapelet-Guided Counterfactual Explanation for Time Series Classification.
\newblock In \emph{2022 IEEE International Conference on Big Data (Big Data)}, 1564--1569. IEEE.

\bibitem[{Li, Boubrahimi, and Hamd(2022)}]{MG-CF}
Li, P.; Boubrahimi, S.~F.; and Hamd, S.~M. 2022.
\newblock Motif-guided Time Series Counterfactual Explanations.
\newblock \emph{arXiv preprint arXiv:2211.04411}.

\bibitem[{Mothilal, Sharma, and Tan(2020)}]{four-measures-0}
Mothilal, R.~K.; Sharma, A.; and Tan, C. 2020.
\newblock Explaining machine learning classifiers through diverse counterfactual explanations.
\newblock In \emph{Proceedings of the 2020 conference on fairness, accountability, and transparency}, 607--617.

\bibitem[{Raschka(2018)}]{mlxtend}
Raschka, S. 2018.
\newblock MLxtend: Providing machine learning and data science utilities and extensions to Python’s scientific computing stack.
\newblock \emph{The Journal of Open Source Software}, 3(24).

\bibitem[{Rudin(2019)}]{need_XAI}
Rudin, C. 2019.
\newblock Stop explaining black box machine learning models for high stakes decisions and use interpretable models instead.
\newblock \emph{Nature machine intelligence}, 1(5): 206--215.

\bibitem[{Sarker(2021{\natexlab{a}})}]{dl_advantage}
Sarker, I.~H. 2021{\natexlab{a}}.
\newblock Deep learning: a comprehensive overview on techniques, taxonomy, applications and research directions.
\newblock \emph{SN Computer Science}, 2(6): 420.

\bibitem[{Sarker(2021{\natexlab{b}})}]{ml_advantage}
Sarker, I.~H. 2021{\natexlab{b}}.
\newblock Machine learning: Algorithms, real-world applications and research directions.
\newblock \emph{SN computer science}, 2(3): 160.

\bibitem[{Schlegel et~al.(2019)Schlegel, Arnout, El-Assady, Oelke, and Keim}]{need_focus_on_ts_data}
Schlegel, U.; Arnout, H.; El-Assady, M.; Oelke, D.; and Keim, D.~A. 2019.
\newblock Towards a rigorous evaluation of XAI methods on time series.
\newblock In \emph{2019 IEEE/CVF International Conference on Computer Vision Workshop (ICCVW)}, 4197--4201. IEEE.

\bibitem[{Theissler et~al.(2022)Theissler, Spinnato, Schlegel, and Guidotti}]{xai_review0}
Theissler, A.; Spinnato, F.; Schlegel, U.; and Guidotti, R. 2022.
\newblock Explainable AI for Time Series Classification: A review, taxonomy and research directions.
\newblock \emph{IEEE Access}.

\bibitem[{Van~Looveren et~al.(2021)Van~Looveren, Klaise, Vacanti, and Cobb}]{conditional_gan_for_all}
Van~Looveren, A.; Klaise, J.; Vacanti, G.; and Cobb, O. 2021.
\newblock Conditional generative models for counterfactual explanations.
\newblock \emph{arXiv preprint arXiv:2101.10123}.

\bibitem[{Wachter, Mittelstadt, and Russell(2017)}]{w-CF}
Wachter, S.; Mittelstadt, B.; and Russell, C. 2017.
\newblock Counterfactual explanations without opening the black box: Automated decisions and the GDPR.
\newblock \emph{Harv. JL \& Tech.}, 31: 841.

\bibitem[{Yoon, Jarrett, and Van~der Schaar(2019)}]{TimeGAN}
Yoon, J.; Jarrett, D.; and Van~der Schaar, M. 2019.
\newblock Time-series generative adversarial networks.
\newblock \emph{Advances in neural information processing systems}, 32.

\end{thebibliography}

\end{document}